\documentclass[cameraready]{Interspeech}
\usepackage{soul}
\usepackage{subcaption}

\title{Preferences of a Voice-First Nation: Large-Scale Pairwise Evaluation and Preference Analysis for TTS in Indian Languages}

\author[affiliation={1,2}, equalcontribution]{Srija}{Anand}
\author[affiliation={1,2}, equalcontribution]{Ashwin}{Sankar}
\author[affiliation={2}]{Ishvinder}{Sethi}
\author[affiliation={3}]{Aaditya}{Pareek}
\author[affiliation={2}]{Kartik}{Rajput}
\author[affiliation={3}]{Gaurav}{Yadav}
\author[affiliation={2}]{Nikhil}{Narasimhan}

\author[affiliation={1}]{Adish}{Pandya}
\author[affiliation={1}]{Deepon}{Halder}
\author[affiliation={1,2}]{Mohammed Safi Ur Rahman}{Khan}

\author[affiliation={1,2}]{Praveen}{S V}
\author[affiliation={3}]{Shobhit}{Banga}
\author[affiliation={1,2}]{Mitesh}{M Khapra}

\address{
    \vspace{-3mm}
  $^1$Indian Institute of Technology Madras, India \quad
  $^2$AI4Bharat, India \quad
  $^3$Josh Talks, India
}

\email{srijaanand@ai4bharat.org, ashwins1211@gmail.com}


\newcommand{\gemini}{{\textsc{Gemini 2.5 Pro TTS}}}
\newcommand{\gpt}{{\textsc{GPT-4o-mini TTS}}}
\newcommand{\eleven}{{\textsc{Eleven Labs v3}}}
\newcommand{\sonic}{{\textsc{Sonic 3}}}
\newcommand{\speech}{{\textsc{Speech 2.8 HD}}}
\newcommand{\bulbul}{{\textsc{Bulbul v3 Beta}}}
\newcommand{\indicf}{{\textsc{Indic F5}}}

\usepackage{xcolor}
\usepackage[normalem]{ulem}
\definecolor{deepBlue}{HTML}{0D47A1}

\usepackage{comment}

\begin{document}

\maketitle
\begin{abstract}
Crowdsourced pairwise evaluation has emerged as a scalable approach for assessing foundation models. However, applying it to Text to Speech(TTS) introduces high variance due to linguistic diversity and multidimensional nature of speech perception. We present a controlled multidimensional pairwise evaluation framework for multilingual TTS that combines linguistic control with perceptually grounded annotation. Using 5K+ native and code-mixed sentences across 10 Indic languages, we evaluate 7 state-of-the-art TTS systems and collect over 120K pairwise comparisons from over 1900 native raters. In addition to overall preference, raters provide judgments across 6 perceptual dimensions: intelligibility, expressiveness, voice quality, liveliness, noise, and hallucinations. Using Bradley-Terry modeling, we construct a multilingual leaderboard, interpret human preference using SHAP analysis and analyze leaderboard reliability alongside model strengths and trade-offs across perceptual dimensions.
%Crowdsourced pairwise evaluation has emerged as a scalable and statistically robust paradigm for assessing foundation models. However, directly applying open-ended pairwise evaluation to Text-to-Speech (TTS) introduces substantial variance due to linguistic diversity and the inherently multidimensional nature of speech perception. We introduce a controlled, multidimensional pairwise evaluation framework for multilingual TTS that combines strict linguistic control with perceptually grounded annotation. Using a curated set of 500+ native and code-mixed sentences per language across 10 Indic languages, we evaluate 7 state-of-the-art TTS systems and collect over 120K pairwise comparisons from 1,500+ native raters. In addition to overall preference, raters provide judgments across six perceptual dimensions: intelligibility, expressiveness, voice quality, liveliness, noise, and hallucinations. Using Bradley--Terry modeling we produce a statistically robust multilingual leaderboard, and using SHAP value analysis, we reveal the perceptual factors driving human preference and the trade-offs underlying model performance. Our results demonstrate the importance of controlled multilingual evaluation and provide both robust ranking and diagnostic insight into the strengths and limitations of modern TTS systems.
\end{abstract}

\section{Introduction}
\begingroup
% \renewcommand{\thefootnote}{}
% \footnotemark\footnotetext{This work is under review at INTERSPEECH 2026}
\endgroup
India is widely recognized as a voice first nation, where many people prefer to access digital services primarily through speech rather than text interfaces. The country’s linguistic diversity, with hundreds of languages and widespread bilingualism, leads to real world speech that frequently includes code-mixing, domain specific vocabulary, numerals, and alphanumeric expressions. Along with the rapid growth of voice driven applications such as education, accessibility, telemedicine, and conversational AI, this has created strong demand for high quality Text to Speech systems in Indian languages. While recent neural TTS advances have improved synthesis quality~\cite{shen2024naturalspeech, ju2024naturalspeech, sankar25_interspeech, chen2025f5ttsfairytalerfakesfluent}, rigorous evaluation remains challenging, and existing studies are often limited in scale, coverage, or diagnostic depth.

Traditional TTS evaluation methods such as MOS \cite{kirkland23_ssw,wester2015are, edlund24_interspeech, dall14_speechprosody}, CMOS \cite{Loizou2011}, and MUSHRA \cite{perrotin2025_blizzard, mushra2015} have long been used to assess perceptual quality, but recent studies \cite{perrotin2025_blizzard, Varadhan2024RethinkingMA, lemaguer2024limits, srinivasavaradhan25_interspeech, anand2024elaichienhancinglowresourcetts, kayyar2023subjective} highlight their limitations for modern multilingual TTS systems. In particular, MOS style evaluations rely on absolute ratings that are sensitive to individual rater calibration~\cite{cooper23_interspeech}. Pairwise preference evaluation \cite{chiang2023report} provides a stronger alternative by capturing direct comparisons and enabling statistically grounded rankings using models such as Bradley-Terry~\cite{bradley1952rank}. However, applying pairwise evaluation to  speech synthesis requires careful control over linguistic variability, and overall preference alone provides limited insight into why raters prefer one system over another. Incorporating fine grained feedback on various perceptual attributes is crucial to reveal the factors driving those human judgments.

To address these challenges, we design a controlled multidimensional pairwise evaluation framework for multilingual TTS with the following contributions. First, we curate a phonetically diverse benchmark consisting of carefully designed evaluation sentences across 10 Indic languages, covering real world linguistic phenomena such as code-mixing, numerals, and deployment realistic expressions. Second, using this benchmark, we conduct a large scale controlled human evaluation of 7 state of the art TTS systems, collecting over 120K pairwise comparisons from more than 1900 native raters across India, along with fine grained feedback across six perceptual dimensions. Third, using Bradley-Terry modeling, we construct a statistically grounded multilingual leaderboard and analyze the perceptual drivers of model preference. Finally, we study evaluation reliability as scale increases, examining the effects of the number of raters, comparisons, and sentences, and provide a fine grained analysis of the perceptual attributes that most strongly influence listener decisions. Together, this study provides a large scale benchmark, systematic insights into reliable evaluation, and practical guidance for designing scalable and interpretable human evaluation frameworks for speech synthesis. The benchmark and the preference data is available at \url{https://huggingface.co/datasets/ai4bharat/SpeechArenaBench/}. We hope this will support future research on multilingual TTS evaluation.

\section{Related Work}

\noindent{\textbf{Subjective evaluation of TTS systems:}}
% Subjective listening tests such as Mean Opinion Score (MOS)~\cite{kirkland23_ssw,wester2015are}, Comparison MOS (CMOS)~\cite{Loizou2011}, and MUSHRA~\cite{mushra2015} remain the standard methods for evaluating TTS quality. However, such studies are often limited in scale, rater diversity, or language coverage, and typically report aggregate scores that do not reveal the perceptual factors driving preferences~\cite{wester2015are, lemaguer2024limits}. Multidimensional extensions such as MUSHRA-DG~\cite{Varadhan2024RethinkingMA} attempt to capture richer perceptual signals but increase evaluation complexity and limit scalability to multiple candidate systems.
Subjective listening tests such as MOS~\cite{kirkland23_ssw,wester2015are}, CMOS~\cite{Loizou2011}, and MUSHRA~\cite{mushra2015} remain standard for evaluating TTS quality. However, these studies are often limited in scale, or language coverage and typically report aggregate scores that obscure the perceptual factors~\cite{wester2015are, lemaguer2024limits}. Multidimensional extensions such as MUSHRA-DG~\cite{Varadhan2024RethinkingMA} capture richer perceptual signals but increase evaluation complexity and limit scalability across multiple systems.

\noindent{\textbf{Pairwise preference and probabilistic ranking:}}
Pairwise preference evaluation~\cite{zhong25c_interspeech} captures relative judgments between samples and enables probabilistic ranking using models such as Bradley-Terry~\cite{bradley1952rank} and Thurstone~\cite{thurstone1927law}. Prior speech synthesis studies~\cite{tts-arena-v2, ArtificialAnalysisTTS2025, CovalTTS2026} have adopted this paradigm but largely focus on overall preference~\cite{srinivasavaradhan25_interspeech}, without incorporating multidimensional perceptual attribution in multilingual settings.

\noindent{\textbf{Evaluation of multilingual and Indic TTS:}}
% Several studies~\cite{Kumar2022Towards, srinivasavaradhan24_interspeech, sankar25_interspeech} evaluate Indic TTS using MOS or CMOS protocols. However, many predate recent advances in neural TTS or remain limited in scale, linguistic diversity, or diagnostic depth. Large-scale multilingual evaluations across multiple Indic languages remain relatively scarce~\cite{Varadhan2024RethinkingMA}.
Several studies~\cite{sankar25_interspeech, Kumar2022Towards, srinivasavaradhan24_interspeech, prakash2023exploring} evaluate Indic TTS using MOS or CMOS protocols, but many predate recent neural TTS advances or remain limited in scale, linguistic diversity, or diagnostic depth. Large-scale multilingual evaluations across Indic languages remain relatively scarce~\cite{Varadhan2024RethinkingMA}.

\section{Evaluation Framework}
We describe our controlled multidimensional pairwise evaluation framework, including the benchmark, rater recruitment, annotation protocol, perceptual axes, and ranking methodology.

\subsection{Benchmark Construction}
\label{sec:bench}

We construct a multilingual evaluation benchmark of 5,357 sentences across 10 Indian languages: Bengali, Gujarati, Hindi, Kannada, Malayalam, Marathi, Odia, Tamil, Telugu, and Urdu. The sentences span 16 deployment relevant domains (Figure~\ref{fig:domain_wise_ranks}) and are collected from public sources following the methodology in ~\cite{srinivasavaradhan24_interspeech}. To probe edge cases, we additionally generate sentences using \textsc{Gemini-3-pro-preview}~\cite{google2025gemini3}, covering stress categories such as tongue twisters, extreme repetitions, and dense STEM content. We also include 100 expressive utterances from RASA-test~\cite{srinivasavaradhan24_interspeech} to evaluate prosody and emotion. Sentences are either collected natively or translated using \textsc{Gemini-3-pro-preview}. \emph{All sentences undergo strict quality assurance by in-house native language experts}, who accept, correct, or discard them based on linguistic accuracy, fluency, and domain-specific terminology.

%We construct a large-scale multilingual evaluation benchmark comprising 5,370 sentences across 10 Indian languages, \textit{viz.}, Bengali, Gujarati, Hindi, Kannada, Malayalam, Marathi, Odia, Tamil, Telugu and Urdu. These sentences span 16 deployment-relevant domains (see x-axis of Figure~\ref{fig:domain_wise_ranks}), designed to reflect real-world usage scenarios of contemporary TTS systems. Following the methodology of~\cite{srinivasavaradhan24_interspeech}, we collect sentences from multiple domains using publicly available sources. To systematically probe edge-case behavior, we additionally generate a subset of sentences using \textsc{Gemini-3-pro-preview}~\cite{google2025gemini3} incorporating controlled linguistic stress categories such as tongue-twisters, extreme word repetitions, and technically dense STEM content. Finally we include 100 expressive utterances from the RASA-test~\cite{srinivasavaradhan24_interspeech} to evaluate prosodic and emotional rendering. For each language, sentences are either collected natively or translated from English using \textsc{Gemini-3-pro-preview}. \emph{All sentences undergo strict quality assurance by in-house native language experts}, who review and accept, correct, or discard them based on linguistic accuracy, fluency, and domain-specific terminology.

\begin{table}[h]
    % \footnotesize
    \fontsize{8pt}{10pt}\selectfont
    \setlength{\tabcolsep}{3pt}
    \centering
    \caption{Statistics of the benchmark and rater demographics.}
    \label{tab:benchmark-stats}
    \begin{tabular}{@{}lr|lr|lr@{}}
    \toprule
    \multicolumn{2}{c|}{\textbf{Benchmark}} & \multicolumn{2}{c|}{\textbf{Raters}} & \multicolumn{2}{c}{\textbf{Age}} \\ \midrule
    \# sentences       & 5357 & \textbf{Total}     & \textbf{1915} & 18--25 & 885 \\
    \# languages       & 10   & Male      & 767  & 25--40 & 916 \\
    \# domains         & 16   & Female    & 1148 & 40--65 & 114 \\
    \# symbolic sent.  & 1752 & \# states & 22   & \multicolumn{2}{c}{\textbf{\# Comparisons}}\\ 
    \# codemixed sent. & 4164 &           &      &   Total  & 120K\\ \bottomrule
    \end{tabular}
\end{table}

The benchmark (Table~\ref{tab:benchmark-stats}) consists of three structured subsets reflecting different input conditions. The \textbf{normalized subset} expands numerals, equations, and acronyms into fully verbalized forms, while the \textbf{symbolic subset} retains numerals, formulas, and operators to test handling of raw text. The \textbf{code-mixed subset} includes intra-sentential English insertions, transliteration-based mixing, and mixed-script sentences, reflecting everyday multilingual usage in India. Utterances span short, medium, and long durations to evaluate both pronunciation accuracy and longer-range prosodic coherence. Audio outputs are generated on demand under consistent conditions, and where supported, multiple voices per model are evaluated. 
% An overview of the benchmark statistics is provided in Table~\ref{tab:benchmark-stats}.

\subsection{Rater Recruitment}

To ensure reliable human evaluations, we implemented a multi-stage rater recruitment and training protocol. Candidates first completed an auditory screening task requiring them to identify the higher-quality sample between a clearly degraded and a clean recording. Those who passed proceeded to a second evaluation where they justified their choices using the perceptual criteria employed in the study (Table~\ref{tab:granular_axes}). Selected raters were then trained on the evaluation guidelines, platform usage, and quality expectations before accessing the annotation interface. All participants provided informed consent, the study protocol underwent internal ethics review, and raters were fairly compensated according to standard industry rates. The demographics of the final rater pool are summarized in Table~\ref{tab:benchmark-stats}.

\subsection{Annotation Protocol}
\label{sec:annotation_protocol}

To mitigate cognitive overload and prevent post-hoc rationalization, we implement a strict two-step pairwise comparison workflow. In the first phase, raters are shown the text prompt and two anonymous, randomized audio samples (\textsc{Model A} and \textsc{Model B}) and must submit a holistic overall preference--\emph{Model A}, \emph{Model B}, \emph{Both Good}, or \emph{Both Bad}--after listening to both samples. Each rater evaluates 150 randomly sampled sentences. Once submitted, this overall choice is permanently locked and the interface for the second phase is revealed, where raters assess the same audio pair across six granular perceptual axes (Table~\ref{tab:granular_axes}) on an identical scale. This ensures that the overall preference reflects the rater’s immediate auditory judgment, while the granular ratings provide an independent diagnostic breakdown. %rather than a post-hoc explanations of the primary vote rather than influencing it. 

\begin{table}[t]
    \centering
    \caption{Granular perceptual axes used for pairwise evaluation.}
    \label{tab:granular_axes}
    \fontsize{8pt}{9pt}\selectfont
    \begin{tabular}{p{2.0cm}|p{5.1cm}}
    \toprule
    \textbf{Axis} & \textbf{Description} \\
    \midrule
    Intelligibility & Clarity and correctness of pronunciation, including native and code-mixed words. \\
    Expressiveness & Appropriateness of prosody, intonation, and emotional delivery. \\
    Voice Quality & Naturalness and human-likeness of the voice, including timbre and absence of artifacts. \\
    Liveliness & Energy and pacing of the speech, indicating engaging vs. monotonous delivery. \\
    Hallucinations & Fidelity to the input text, penalizing skipped words or unintended sounds. \\
    Presence of Noise & Background artifacts or signal distortions such as hiss, clicks, or static. \\
    \bottomrule
    \end{tabular}
\end{table}

% \subsection{Models Evaluated}
% \label{sec:models}

% We evaluate seven state-of-the-art TTS systems spanning commercial production APIs, open-source models and Indic-specialized models. These include \textit{Gemini 2.5 Pro TTS and GPT-4o Mini TTS}, which represent multimodal foundation models with integrated speech generation; \textit{Eleven Labs v3, Sonic 3, and Speech 2.8 HD}, which are commercial neural TTS systems designed for high-quality and expressive speech synthesis; \textit{Bulbul V3 Beta}, a regional model optimized for Indic languages and code-mixed speech; and \textit{Indic F5}, an open-source multilingual TTS model based on the F5-TTS architecture. Together, these systems represent diverse architectural paradigms and deployment settings, enabling comprehensive evaluation of multilingual speech synthesis performance.

\subsection{Ranking and Statistical Modelling}
We convert the collected pairwise preferences into a single leaderboard by fitting a maximum-likelihood Bradley–-Terry (BT) model~\cite{bradley1952rank, hunter2004_mmbt}. The resulting latent scores are mapped onto an Elo-like scale for consistency with standard leaderboard conventions~\cite{chiang2024chatbot}. To quantify uncertainty, we perform bootstrap resampling~\cite{diciccio1996_bootstrapconfidence} of the preference data 500 times and refit the BT model on each sample to obtain $95\%$ confidence intervals. For ranking, we use a conservative, significance-aware criterion: a system is  strictly better than another only if its confidence interval lies entirely above that of the other system~\cite{chiang2024chatbot}.

\section{Results}
We evaluate 7 state-of-the-art TTS systems---\gemini, \gpt, \eleven, \sonic, \speech, \bulbul, and \indicf~\cite{varadhan2025phirherafairyenglish}---spanning commercial production APIs, open-source systems, and Indic-specialized models. To ensure fair comparison across systems, all models were evaluated using identical text prompts without style conditioning. Each system used its recommended default voice configuration, and audio samples were generated in non-streaming mode under consistent inference settings. When multiple voices were available, we ensured that pairwise comparisons were performed between voices of the same gender to avoid confounding effects due to gender-related perceptual differences. In this section, we present the multilingual leaderboards and provide detailed analyses of ranking reliability, perceptual drivers of human preference, acoustic trade-offs between models, and the impact of code-mixed inputs.

% \begin{table}[t]
% \fontsize{7pt}{9pt}\selectfont
% \setlength{\tabcolsep}{3pt}
% \centering
% \caption{Multilingual pairwise leaderboard based on Bradley--Terry modeling. Confidence intervals (95\%) are shown as $\pm$.}
% \label{tab:leaderboard}
% \begin{tabular}{ccrrcr}
% \toprule
% \textbf{Rank} & \textbf{Model} & \multicolumn{1}{c}{{\begin{tabular}[c]{@{}c@{}}\textbf{Score} $\pm$\\ 
% \textbf{95\% CIs}\end{tabular}}} & \multicolumn{1}{c}{{\begin{tabular}[c]{@{}c@{}}\textbf{\#} \\ \textbf{battles}\end{tabular}}} & \multicolumn{1}{c}{{\begin{tabular}[c]{@{}c@{}}\textbf{Win}\\ \textbf{Rate (\%)}\end{tabular}}} & \multicolumn{1}{c}{{\begin{tabular}[c]{@{}c@{}}\textbf{\#} \\ \textbf{lang}\end{tabular}}} \\
% \midrule
% 1 & Gemini 2.5 Pro TTS & $1126.46 \pm 3.00$ & 39,835 & 70 & 10 \\
% 2 & Eleven Labs v3 & $1057.25 \pm 3.00$ & 30,044 & 57 & 9 \\
% 2 & Sonic 3 & $1053.46 \pm 3.00$ & 30,952 & 56 & 8 \\
% 4 & Bulbul V3 Beta & $1020.56 \pm 3.00$ & 37,405 & 51 & 9 \\
% 5 & Speech 2.8 HD & $991.35 \pm 6.00$ & 7,013 & 48 & 2 \\
% 6 & GPT 4o Mini TTS & $929.00 \pm 5.00$ & 12,978 & 38 & 5 \\
% 7 & Indic F5 & $821.92 \pm 3.00$ & 35,823 & 21 & 10 \\
% \bottomrule
% \end{tabular}
% \end{table}

% \subsection{Multilingual Leaderboard}
% \textbf{RQ1. How do modern TTS systems rank overall and within each language?}

% \subsection{How do modern TTS systems rank overall and within each language?}

\begin{table}[h]
\fontsize{8pt}{9pt}\selectfont
\setlength{\tabcolsep}{3pt}
\centering
\caption{Overall leaderboard based on Bradley-Terry scores ($\uparrow$). Win rate is reported as a percentage. \# comp is the number of comparisons. \# lang is the number of supported languages.}
\label{tab:leaderboard}
\begin{tabular}{ccrrcr}
\toprule
\textbf{Rank} & \textbf{Model} & \multicolumn{1}{c}{{\begin{tabular}[c]{@{}c@{}}\textbf{Score} $\pm$\\ 
\textbf{95\% CIs}\end{tabular}}} & \multicolumn{1}{c}{{\begin{tabular}[c]{@{}c@{}}\textbf{\#} \\ \textbf{comp}\end{tabular}}} & \multicolumn{1}{c}{{\begin{tabular}[c]{@{}c@{}}\textbf{Win}\\ \textbf{Rate}\end{tabular}}} & \multicolumn{1}{c}{{\begin{tabular}[c]{@{}c@{}}\textbf{\#} \\ \textbf{lang}\end{tabular}}} \\
\midrule
1 & \gemini & $1128.53 \pm 3$ & 46,023 & 70 & 10 \\
2 & \eleven & $1056.28 \pm 2$ & 40,800 & 57 & 9 \\
2 & \sonic & $1050.83 \pm 3$ & 33,795 & 56 & 8 \\
4 & \bulbul & $1021.91 \pm 3$ & 42,221 & 52 & 9 \\
5 & \speech & $993.94 \pm 6$ & 7,834 & 47 & 2 \\
6 & \gpt & $942.76 \pm 4$ & 15,207 & 40 & 5 \\
7 & \indicf & $805.75 \pm 3$ & 42,130 & 19 & 10 \\
\bottomrule
\end{tabular}
\end{table}

% \noindent \textbf{Overall Scores.}
% Table~\ref{tab:leaderboard} presents the multilingual leaderboard constructed using Bradley--Terry modeling over 120K pairwise comparisons. \textsc{Gemini 2.5 Pro TTS} ranks first with a score of $1128.53 \pm 3$ and a 70\% win rate, establishing a clear margin over competing systems. Eleven Labs v3 ($1056.28 \pm 2$, 57\%) and Sonic 3 ($1050.83 \pm 3$, 56\%) follow with closely matched performance and are statistically indistinguishable under bootstrap uncertainty. In contrast, the open-source Indic F5 model ranks last despite extensive evaluation coverage (10 languages and 42K battles), with a score of $805.75 \pm 3$ and a win rate of 19\%, indicating a substantial perceptual gap relative to the commercial systems evaluated in this study. Across most adjacent systems, score separations exceed their bootstrap confidence intervals, indicating that the evaluation scale is sufficient to resolve the majority of performance differences. When intervals overlap, the Bradley--Terry model assigns shared ranks, reflecting statistical indistinguishability rather than imposing an arbitrary ordering. We next analyze language-wise leaderboards to assess whether the aggregate ordering reflects consistent multilingual performance or is driven by a subset of languages.
\subsection{Overall and Language-wise Rankings of TTS Systems}
\begin{figure}[h]
    \centering
    \includegraphics[width=1\linewidth]{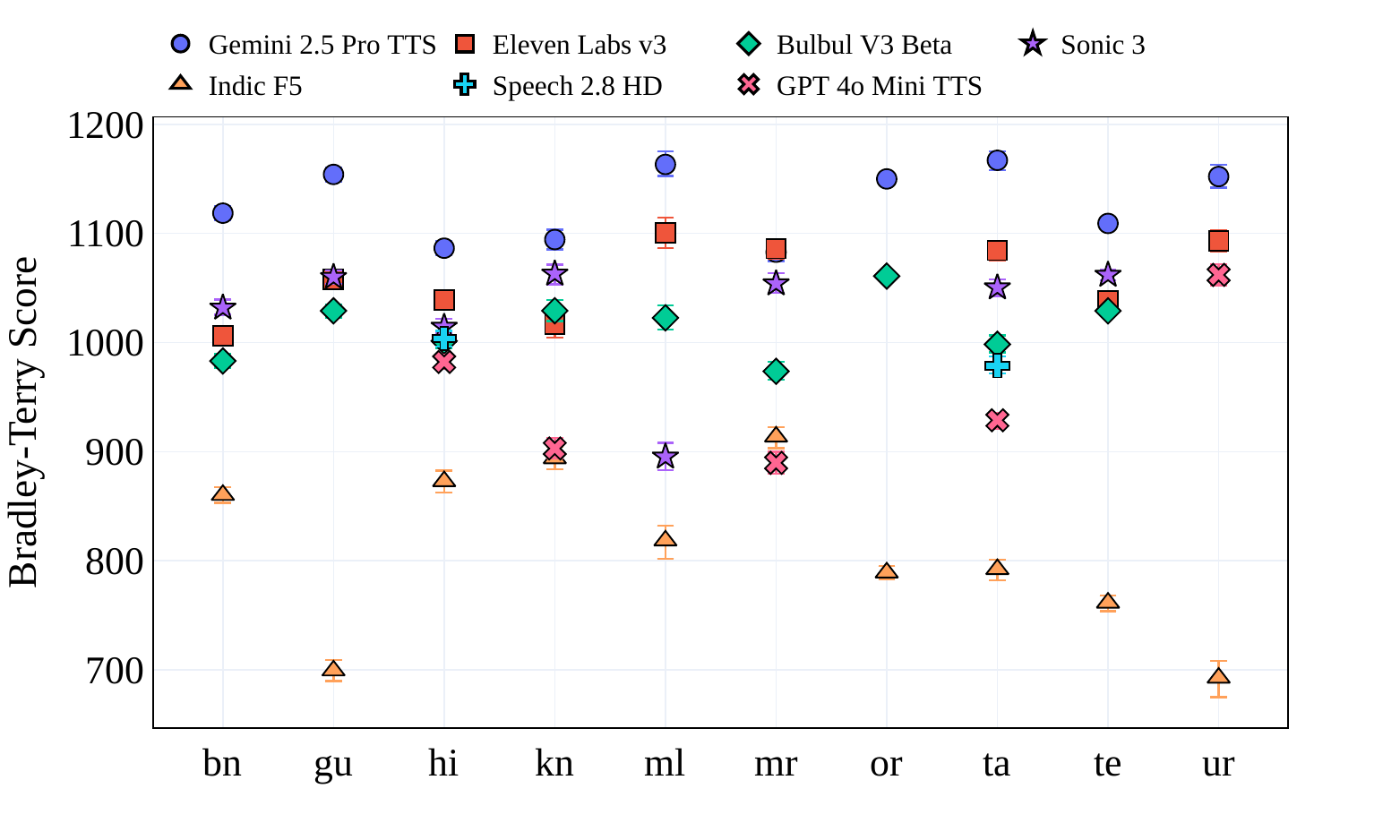}
    \caption{Language-wise rankings of evaluated TTS systems based on Bradley–Terry (BT) scores with bootstrap confidence intervals. Languages are represented using ISO-639-2 codes.}
    \label{fig:langwise_leaderboard}
\end{figure}

Table~\ref{tab:leaderboard} shows the overall leaderboard based on over 120K pairwise comparisons. \gemini~ranks first ($1128.53 \pm 3$), followed by \eleven~($1056.28 \pm 2$) and \sonic~($1050.83 \pm 3$), which are statistically indistinguishable. In contrast, the open-source \indicf~model ranks last ($805.75 \pm 3$) despite extensive evaluation coverage (10 languages and 42K pairwise comparisons) indicating a large gap relative to the other commercial systems evaluated. Most adjacent score differences exceed their bootstrap confidence intervals, indicating that the evaluation scale is sufficiently sensitive to resolve the majority of performance differences.% We next analyze language-wise leaderboards to assess whether the aggregate ordering reflects consistent multilingual performance or is driven by a subset of languages.
Figure~\ref{fig:langwise_leaderboard} presents per-language rankings. \gemini~ranks first in 9 of 10 languages, with near parity with \eleven~in the case of Marathi. Rankings among \eleven, \sonic, and \bulbul~vary across languages with relatively small differences while \indicf~consistently ranks at or near the bottom.

\subsection{Understanding Sensitivity of Leaderboard Rankings}

\begin{figure}[h]
    \centering
    \includegraphics[width=1\linewidth]{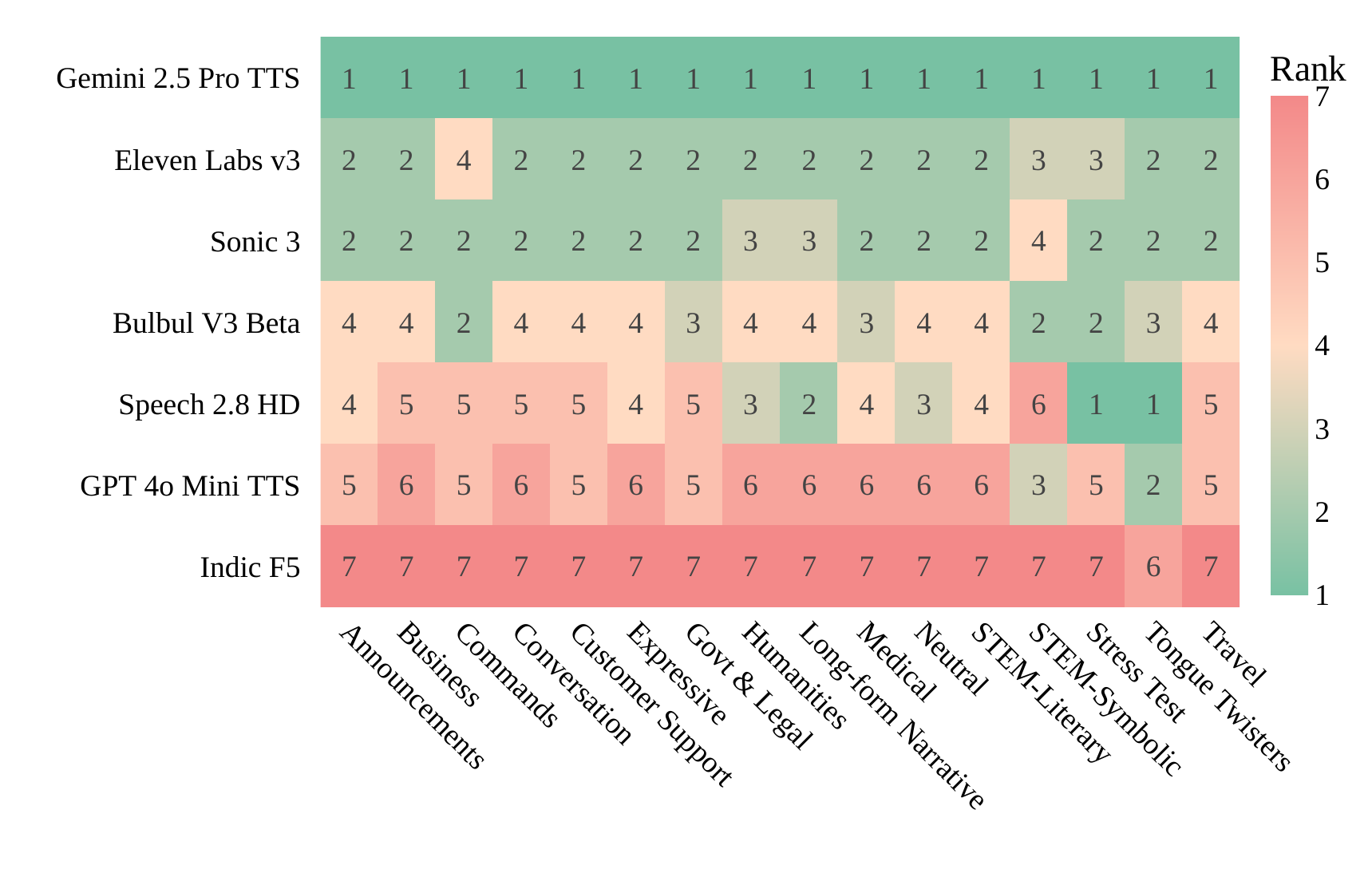}
    \caption{System ranks shift across benchmark domains.}
    
    \label{fig:domain_wise_ranks}
\end{figure}

\noindent \textbf{How Do Ranks Vary Across Domains?}
Figure~\ref{fig:domain_wise_ranks} shows model rankings across domains. \gemini~ranks first in all 16 domains, indicating consistent strong performance. In contrast, rankings among \eleven, \sonic, and \bulbul~vary by domain, showing smaller performance differences. Some domains diverge from the aggregate ordering: \speech~ranks first in the Stress Test category, and multiple systems tie in Tongue Twisters. \indicf~remains near the bottom across domains. Overall, domain-wise analysis reveals meaningful variation among competing systems~\cite{edlund24_interspeech}.

% \noindent \textbf{RQ2. How sensitive are leaderboard rankings to input conditions?}
% We next examine whether leaderboard positions remain stable when conditioning on input type: \emph{Native} (fully normalized text in the target language), \emph{Symbolic} (containing numerals, abbreviations, and mathematical operators), and \emph{Code-mixed} (mixed English–Indic content). Table~\ref{tab:options} reports aggregate Bradley--Terry ratings computed independently for each input condition. At the top of the leaderboard, \textbf{\gemini} remains first across all three conditions, indicating strong robustness to input variation.
% Overall, the rankings change only modestly across subsets. Nevertheless, some condition-specific differences are visible. For example, \bulbul improves under symbolic inputs relative to native text. These subsets capture challenges that arise in real-world usage. Code-mixed inputs reflect common multilingual text in India, while symbolic inputs introduce numerals and structured expressions that require reliable normalization during synthesis.  Although the resulting rank shifts are modest, these subsets expose different evaluation challenges. These results highlight the importance of evaluating TTS systems under input conditions that reflect their intended deployment.

\noindent \textbf{How Do Rankings Change with Input Type?}
We examine leaderboard stability across the three subsets discussed in $\S$\ref{sec:bench}: \emph{Normalized}, \emph{Symbolic}, and \emph{Code-mixed}. Table~\ref{tab:options} reports BT ratings computed separately for each subset. \gemini~ranks first under all conditions, indicating robustness to input variation. Overall, rankings change only modestly, though some shifts are observed - for example, \bulbul~performs relatively better under symbolic inputs. These subsets reflect real-world deployment challenges, including normalization of structured expressions and multilingual text. While rank differences are limited, condition-wise evaluation reveals complementary strengths and weaknesses across systems.

\begin{table}[h]
\centering
\fontsize{8pt}{9pt}\selectfont
\setlength{\tabcolsep}{3pt}
\caption{Bradley–Terry scores ($\uparrow$) across benchmark input types: Code-mixed, Normalized and Symbolic.}
\begin{tabular}{@{}lrrr@{}}
\toprule
                   & \textbf{Codemixed} & \textbf{Normalized}  & \textbf{Symbolic} \\ \midrule
\gemini & $1135.45 \pm 3$   & $1120.12 \pm 3$ & $1143.68 \pm 5$  \\
\eleven     & $1054.00 \pm 3$  & $1059.28 \pm 3$ & $1044.37 \pm 5$ \\
\sonic           & $1054.74 \pm 3$  & $1049.68 \pm 3$ & $1049.42  \pm 6$\\
\bulbul     & $1031.28 \pm 3$  & $1012.58  \pm 3$ & $1048.20  \pm 5$\\
\speech     & $982.76  \pm 7$  & $1011.02 \pm 6$ & $958.15   \pm 10$ \\
\gpt    & $951.42  \pm 5$  & $934.76  \pm 5$& $970.75   \pm 8$ \\
\indicf          & $812.54  \pm 4$  & $849.75  \pm 4$ & $785.42   \pm 6$ \\ \bottomrule
\end{tabular}
\label{tab:options}
\end{table}

% \noindent \strikeadd{\textbf{RQ4. How do systems rank across individual perceptual dimensions?}}{}

% \textit{Acoustic trade-offs between perceptual attributes}
% Importantly, we do not observe strong antagonistic trade-offs between expressiveness and robustness; instead, models largely differ in overall performance level and balance across dimensions. This suggests that improvements in expressiveness need not inherently increase hallucination or noise, and that high-quality synthesis can simultaneously achieve perceptual richness and acoustic fidelity.

\subsection{Interpreting Human Preference}
\begin{figure}[h]
    \centering
    \includegraphics[width=\linewidth]{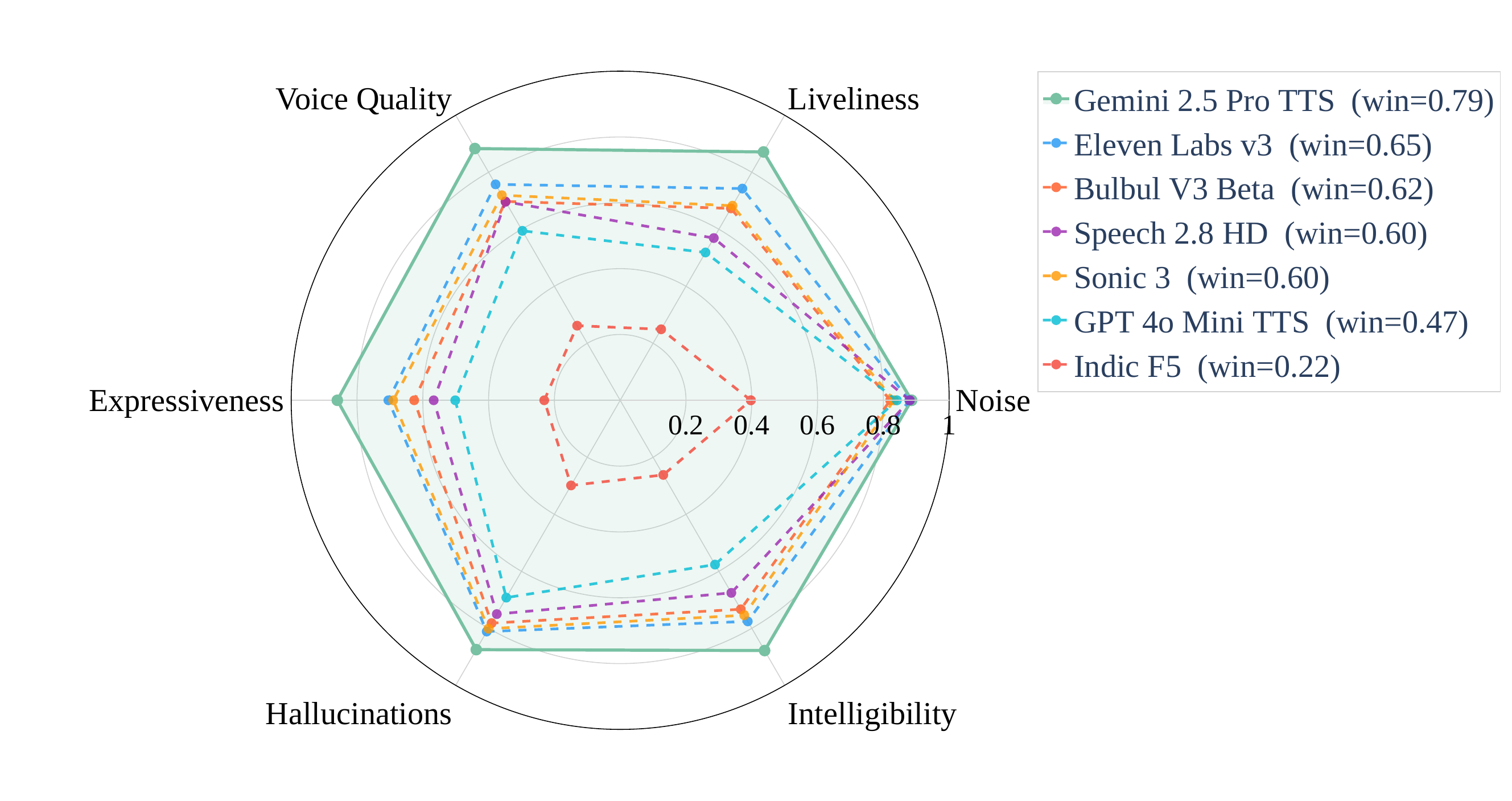}
    % \caption{Perceptual performance profiles of evaluated TTS systems across the six granular axes. Values represent average pairwise win rates for each perceptual dimension.}
    \caption{Multi-dimensional perceptual performance of TTS systems measured by average win rates across six axes.}
    \label{fig:win_rates_perceptual_axes}
\end{figure}

\noindent \textbf{What Do Perceptual Axes Reveal About Human Preference?}
To better understand how raters form preferences, we analyze performance across six perceptual axes highlighted in Table~\ref{tab:granular_axes} using average axis-level win rates (Figure~\ref{fig:win_rates_perceptual_axes}). Higher scores on the hallucination and noise axes correspond to fewer artifacts and cleaner audio, and therefore indicate stronger robustness. \gemini~performs consistently well on all axes, reflecting balanced strengths in expressiveness, intelligibility, liveliness, voice quality, and robustness. \eleven, \sonic, and \bulbul~maintain strong intelligibility and robustness but exhibit comparatively lower expressiveness and liveliness. \speech~and \gpt~show more uneven performance across dimensions, while \indicf~ranks lower on most perceptual axes.

% To move beyond scalar leaderboard scores, we analyze model performance across the six perceptual axes using average axis-level win rates (Figure~\ref{fig:win_rates_perceptual_axes}). Higher scores on hallucinations and noise correspond to fewer hallucinated tokens and cleaner audio, respectively, and therefore indicate stronger robustness.
% The radar visualization reveals clear stratification across models. \gemini~achieves consistently high scores across all dimensions, indicating balanced performance in expressiveness, intelligibility, liveliness, voice quality, and robustness (low hallucination and noise rates). \eleven, \sonic, and \bulbul~exhibit strong robustness and intelligibility, with moderate reductions in expressiveness and liveliness relative to the top model. \speech~shows comparatively balanced but lower overall performance. \gpt~demonstrates moderate intelligibility and noise robustness but reduced performance in expressiveness and liveliness. Indic F5 trails across most perceptual dimensions. These results demonstrate that overall ranking alone obscures meaningful structural differences in perceptual balance. Multidimensional analysis provides a clearer understanding of how models distribute performance across expressive and robustness-related attributes.

\noindent \textbf{Can Granular Judgments Predict Overall Preference?}
% In the preceding sections, overall preference was treated as a scalar outcome summarizing pairwise comparisons. While this establishes a statistically grounded ranking, it leaves open the question of how granular perceptual judgments combine to produce holistic preferences. We therefore investigate whether overall preference can be reconstructed from these finer-grained evaluations. For each pairwise comparison, we construct a binary feature vector indicating whether Model~A performs at least as well as Model~B on each perceptual axis. The feature value is set to 1 if Model~A is judged better or tied-good on that axis, and 0 if Model~A loses or the comparison is marked as both-bad. The task is then to predict whether Model~A is the overall preferred system. We train a gradient-boosted decision tree classifier (XGBoost) using these six features. The model uses 300 trees with maximum depth 4 and learning rate 0.05. To evaluate generalization, we train the classifier on high-resource languages and evaluate it on held-out languages. 
Overall preference provides a reliable ranking, but it does not reveal how raters combine multiple perceptual cues into a single judgment. We therefore test whether overall preference can be reconstructed from granular axis-level evaluations. For each comparison, we construct a binary feature vector indicating whether Model~A performs at least as well as Model~B on each axis (1 = better or both-good; 0 = worse or both-bad), and train an XGBoost~\cite{chen2016_xgboost} classifier to predict the preferred system (label = 1, if Model~A was the preferred system, 0 otherwise). We train the model on a subset of languages and evaluate it on held-out languages (Bengali, Kannada, Malayalam, Marathi, \& Urdu), achieving 86.1\% accuracy with consistent performance across languages (83.6\%–91.0\%). \emph{This suggests that raters rely on stable and transferable perceptual criteria when forming overall judgments across linguistic settings.} 

\begin{figure}[h]
    \centering
    \includegraphics[width=0.95\linewidth]{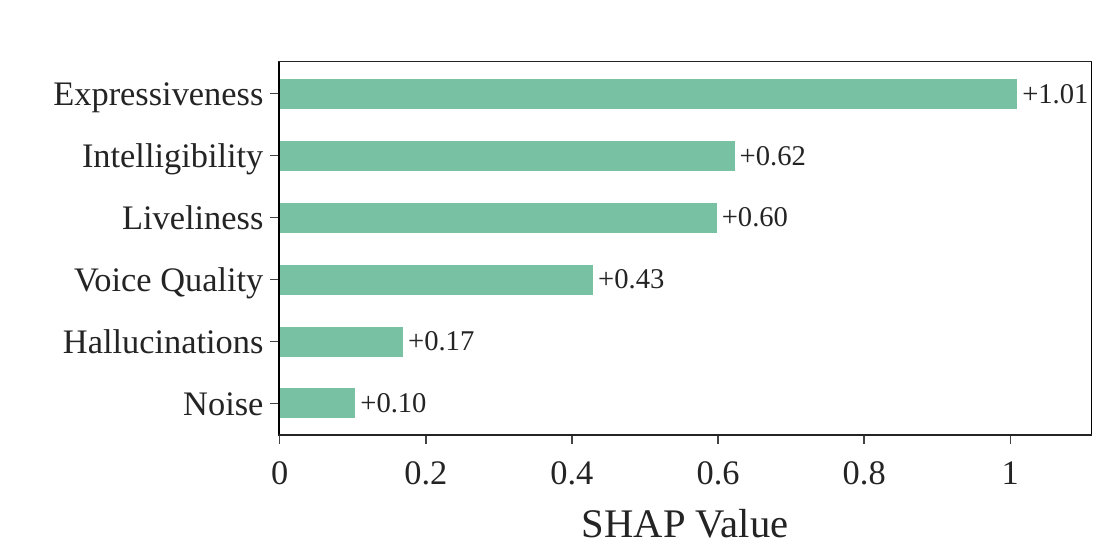}
    % \caption{SHAP feature importance for the preference prediction model. Each point represents a pairwise comparison, with position indicating its contribution to the predicted winner. Color denotes the feature value (blue: low, red: high). Features are ordered by mean absolute SHAP value.}SHAP feature importance for the preference prediction model under the language holdout setting.
    \caption{Mean absolute SHAP values showing the relative contribution of each perceptual axis to overall preference.}
    \label{fig:holdout_beeswarm}
\end{figure}

\noindent \textbf{Which Axes Drive Preference?}
To understand which perceptual attributes most strongly influence listener preference, we perform SHAP~\cite{lundberg2017_shape} (SHapley Additive exPlanations) analysis (Figure~\ref{fig:holdout_beeswarm}) on the trained model. The results show that \textit{expressiveness} and \textit{intelligibility} are the strongest predictors of overall preference, followed by \textit{liveliness} and \textit{voice quality}. In contrast, \textit{hallucinations} and \textit{noise} contribute less to the model's predictions. This does not imply that robustness is unimportant; rather, most evaluated systems already achieve relatively strong performance on these axes, leaving limited variation to differentiate models. \emph{Overall, these results suggest that rater preference is primarily driven by expressive and intelligible speech once basic robustness to noise and hallucinations is satisfied.}

\subsection{When Does the Leaderboard Become Reliable?}

% \begin{figure}[h]
%     \centering
%     \includegraphics[width=1\linewidth]{figures/rater_sentence_2x2_grid.png}
%     \caption{Leaderboard reliability vs. evaluation scale. Rank consistency (Spearman’s $\rho$) and Bradley–-Terry uncertainty as raters and sentences increase.}
%     \label{fig:rank-consistency-uncertainity}
% \end{figure}

\begin{figure}[h]
    \centering
    \includegraphics[width=1\linewidth]{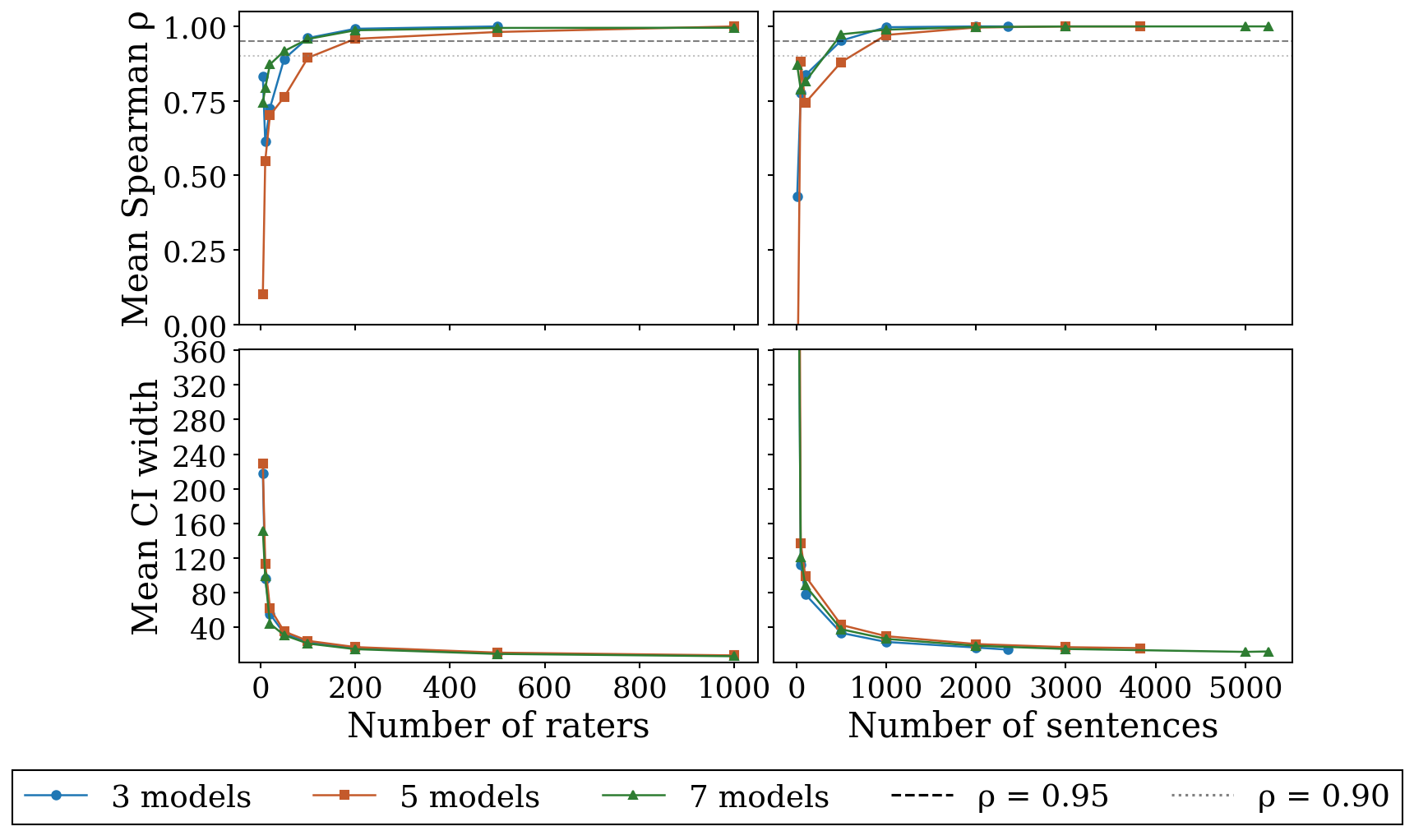}
    \caption{Rank consistency (Spearman’s $\rho$) and BT uncertainty as the number of raters increases (left) and as the number of sentences increases with 200 raters fixed (right).}
    %Rankings stabilize near 200 raters ($\rho \approx 0.95$), while more sentences primarily reduce score uncertainty.}
    % \caption{Left: Rank consistency (Spearman’s $\rho$) and Bradley–-Terry uncertainty as the number of raters increase, measured for three setups. Around 200 raters yield $\rho \approx 0.95$, indicating stable rankings. Right: With 200 raters fixed, we vary the number of evaluation sentences to find the minimum number of sentences required to achieve rank consistency.}
    \label{fig:rank-consistency-uncertainity}
\end{figure}

% A central practical question in large-scale human evaluation is \emph{sample efficiency}: how many judgments are needed before the ranking stops changing in meaningful ways. We quantify leaderboard reliability as we increase (i) the number of raters and (ii) the number of unique sentences, reporting two complementary signals: (1) \emph{rank consistency} measured by Spearman’s $\rho$ against the full-evaluation leaderboard, and (2) \emph{statistical uncertainty} measured by the mean width of 95\% bootstrap confidence intervals (CIs) on Bradley--Terry scores.
A central question in large-scale human evaluation is \emph{sample efficiency}: how many judgments are required before leaderboard rankings stabilize. We measure reliability as the number of raters and sentences increases using two signals: (i) rank consistency (Spearman’s $\rho$) with respect to the full-evaluation leaderboard, and (ii) statistical uncertainty measured by the mean width of 95\% bootstrap confidence intervals.
%on BT scores.

% \noindent \textbf{How many raters are enough?}
% Using $\rho \ge 0.95$ as a reliability target, we find in Figure \ref{fig:rank-consistency-uncertainity} that stable rankings typically emerge with 100--200 raters, depending on the number of systems being compared. For example, with 5 systems, 200 raters are required to exceed $\rho \ge 0.95$, yielding a mean CI width of 17.36. With 7 systems, $\rho \ge 0.95$ is reached at 100 raters (mean CI width 21.47).  At these scales the ordering of systems is largely stable, although the confidence intervals around their scores remain non-trivial. This means the leaderboard ranking is reliable, but small score differences between nearby systems should still be interpreted cautiously. \todo{needs discussion}

% \noindent \textbf{How many sentences are enough?}
% We next fix the number of raters at 200 and vary the number of unique sentences. Because the benchmark spans multiple languages, sentence sampling is stratified to ensure balanced coverage across languages. Using the same reliability target ($\rho \ge 0.95$), we find that approximately 1000 sentences are required to achieve stable rankings for setups with 5 and 7 systems and approximately 500 sentences are required to achieve stable rankings for the setup with 3 systems. Increasing the number of sentences beyond this point further reduces score uncertainty but yields diminishing returns in rank consistency.

\noindent \textbf{How Many Raters Are Enough?}
Using $\rho \ge 0.95$ as a reliability target, Figure~\ref{fig:rank-consistency-uncertainity} (left half) shows that stable rankings typically emerge with 100-200 raters, depending on the number of systems compared. For example, with 5 systems, approximately 200 raters are required to exceed $\rho \ge 0.95$, yielding a $\mu_{CI}$ of 17.36. With 7 systems, the same threshold is reached at around 100 raters ($\mu_{CI}=21.47$). At these scales the ordering of systems is largely stable, although confidence intervals around their scores remain non-trivial. This indicates that rank stability is achieved earlier than precise score estimation: the leaderboard ordering is reliable, but small score differences between nearby systems should be interpreted cautiously. %\todo{needs discussion}

\noindent \textbf{How Many Sentences Are Enough?}
Next, fixing the number of raters at 200, we vary the number of unique sentences (Figure ~\ref{fig:rank-consistency-uncertainity} , right half). Because the benchmark spans multiple languages, sentence sampling is stratified to ensure balanced coverage across languages. Using the same reliability criterion ($\rho \ge 0.95$), we find that approximately 1000 sentences are required to achieve stable rankings for setups with 5 and 7 systems, while about 500 sentences suffice for setups with 3 systems. Increasing the number of sentences beyond this point primarily reduces score uncertainty while providing limited additional gains in rank stability.

\section{Conclusion}
We present a controlled, multidimensional pairwise evaluation framework for multilingual TTS systems. Using 5.3K sentences across 10 Indic languages, we collect over 120K pairwise judgments from 1900+ vetted native raters and construct a leaderboard with Bradley-Terry modeling. Beyond aggregate rankings, our six perceptual axes support fine grained diagnostic analysis, with axis-level judgments strongly predicting overall preference across languages, while SHAP analysis shows that once basic robustness to noise and hallucinations is ensured, expressiveness and intelligibility drive listener choice. Our reliability study further demonstrates that stable rankings can be obtained with moderate rater counts when sentence coverage is sufficiently broad. We hope this contribution enables more reliable and interpretable evaluation of multilingual TTS systems.

\ifcameraready
\section{Acknowledgments}
We are grateful to the EkStep Foundation and Nilekani Philanthropies for their generous support, which enabled the recruitment of human resources and provided access to the cloud infrastructure essential for carrying out this work. We thank Google for supporting Praveen and Safi through the Google Ph.D. Fellowship. We extend our heartfelt appreciation to all participants who generously contributed their time to this effort, without whom this work would not have been possible. We also thank the language experts who contributed to benchmark construction and verification, thereby enabling the successful completion of this project. Finally, we acknowledge everyone who contributed to this initiative in any capacity and helped make this work meaningful and impactful.
\fi

\section{Generative AI Use Disclosure}
Generative AI tools were used solely for language polishing and editing during the preparation of this manuscript. These tools assisted with improving clarity, grammar, and conciseness of the writing. No generative AI system was used to generate experimental results, analyses, figures, or scientific conclusions. All technical content, experiments, and interpretations were developed and verified by the authors.

\bibliographystyle{IEEEtran}
\bibliography{mybib}

@article{Varadhan2024RethinkingMA,
  title={Rethinking MUSHRA: Addressing Modern Challenges in Text-to-Speech Evaluation},
  author={Praveena Varadhan and Amogh Gulati and Ashwin Sankar and Srija Anand and Anirudh Gupta and Anirudh Mukherjee and Shiva Kumar Marepally and Ankur Bhatia and Saloni Jaju and Suvrat Bhooshan and Mitesh M. Khapra},
  journal={Trans. Mach. Learn. Res.},
  year={2024},
  volume={2025},
  url={https://api.semanticscholar.org/CorpusID:274141640}
}

@misc{chiang2024chatbot,
    title={Chatbot Arena: An Open Platform for Evaluating LLMs by Human Preference},
    author={Wei-Lin Chiang and Lianmin Zheng and Ying Sheng and Anastasios Nikolas Angelopoulos and Tianle Li and Dacheng Li and Hao Zhang and Banghua Zhu and Michael Jordan and Joseph E. Gonzalez and Ion Stoica},
    year={2024},
    eprint={2403.04132},
    archivePrefix={arXiv},
    primaryClass={cs.AI}
}

@online{google2025gemini3,
  title        = {A New Era of Intelligence with Gemini 3},
  author       = {{Google}},
  year         = {2025},
  url          = {https://blog.google/products-and-platforms/products/gemini/gemini-3/},
  note         = {Google Blog, accessed: 2026-03-04}
}

@article{bradley1952rank,
 ISSN = {00063444, 14643510},
 URL = {http://www.jstor.org/stable/2334029},
 author = {Ralph Allan Bradley and Milton E. Terry},
 journal = {Biometrika},
 number = {3/4},
 pages = {324--345},
 publisher = {[Oxford University Press, Biometrika Trust]},
 title = {Rank Analysis of Incomplete Block Designs: I. The Method of Paired Comparisons},
 urldate = {2026-03-02},
 volume = {39},
 year = {1952}
}

@article{thurstone1927law,
  title={A law of comparative judgment.},
  author={Thurstone, L. L.},
  journal={Psychological Review},
  volume={34},
  number={4},
  pages={273--286},
  year={1927},
  publisher={Psychological Review Company},
  doi={10.1037/h0070288}
}

@misc{tts-arena-v2,
        title        = {TTS Arena 2.0: Benchmarking Text-to-Speech Models in the Wild},
        author       = {mrfakename and Srivastav, Vaibhav and Fourrier, Clémentine and Pouget, Lucain and Lacombe, Yoach and main and Gandhi, Sanchit and Passos, Apolinário and Cuenca, Pedro},
        year         = 2025,
        publisher    = {Hugging Face},
        howpublished = "\url{https://huggingface.co/spaces/TTS-AGI/TTS-Arena-V2}"
}

@misc{ArtificialAnalysisTTS2025,
  author       = {{Artificial Analysis}},
  title        = {Text-to-Speech Leaderboard},
  year         = {2025},
  url          = {https://artificialanalysis.ai/text-to-speech/leaderboard},
  note         = {Accessed: 2025-03-04},
  howpublished = {ArtificialAnalysis.ai}
}

@misc{CovalTTS2026,
  author       = {Coval},
  title        = {{TTS} Benchmarks: Evaluating Latency and Quality of Text-to-Speech Models},
  year         = {2026},
  url          = {https://app.coval.dev/tts-benchmarks},
  note         = {Accessed: 2026-03-04},
  howpublished = {Coval.dev}
}

@misc{anand2024elaichienhancinglowresourcetts,
      title={ELAICHI: Enhancing Low-resource TTS by Addressing Infrequent and Low-frequency Character Bigrams}, 
      author={Srija Anand and Praveen Srinivasa Varadhan and Mehak Singal and Mitesh M. Khapra},
      year={2024},
      eprint={2410.17901},
      archivePrefix={arXiv},
      primaryClass={cs.CL},
      url={https://arxiv.org/abs/2410.17901}, 
}

@inproceedings{srinivasavaradhan25_interspeech,
  title     = {{The State Of TTS: A Case Study with Human Fooling Rates}},
  author    = {Praveen {Srinivasa Varadhan} and Sherry Thomas and Sai {Teja M S} and Suvrat Bhooshan and Mitesh M. Khapra},
  year      = {2025},
  booktitle = {{Interspeech 2025}},
  pages     = {2285--2289},
  doi       = {10.21437/Interspeech.2025-2765},
  issn      = {2958-1796},
}

@inproceedings{Kumar2022Towards,
  author={Kumar, Gokul Karthik and S V, Praveen and Kumar, Pratyush and Khapra, Mitesh M. and Nandakumar, Karthik},
  booktitle={ICASSP 2023 - 2023 IEEE International Conference on Acoustics, Speech and Signal Processing (ICASSP)}, 
  title={Towards Building Text-to-Speech Systems for the Next Billion Users}, 
  year={2023},
  volume={},
  number={},
  pages={1-5},
  doi={10.1109/ICASSP49357.2023.10096069}}

@inproceedings{srinivasavaradhan24_interspeech,
  title     = {{Rasa: Building Expressive Speech Synthesis Systems for Indian Languages in Low-resource Settings}},
  author    = {Praveen {Srinivasa Varadhan} and Ashwin Sankar and Giri Raju and Mitesh M Khapra},
  year      = {2024},
  booktitle = {{Interspeech 2024}},
  pages     = {1830--1834},
  doi       = {10.21437/Interspeech.2024-2421},
  issn      = {2958-1796},
}

@inproceedings{sankar25_interspeech,
  title     = {{Rasmalai : Resources for Adaptive Speech Modeling in IndiAn Languages with Accents and Intonations}},
  author    = {Ashwin Sankar and Yoach Lacombe and Sherry Thomas and Praveen {Srinivasa Varadhan} and Sanchit Gandhi and Mitesh M. Khapra},
  year      = {2025},
  booktitle = {{Interspeech 2025}},
  pages     = {4128--4132},
  doi       = {10.21437/Interspeech.2025-2758},
  issn      = {2958-1796},
}

@inproceedings{edlund24_interspeech,
  title     = {{Assessing the impact of contextual framing on subjective TTS quality}},
  author    = {Jens Edlund and Christina Tånnander and Sébastien {Le Maguer} and Petra Wagner},
  year      = {2024},
  booktitle = {{Interspeech 2024}},
  pages     = {1205--1209},
  doi       = {10.21437/Interspeech.2024-781},
  issn      = {2958-1796},
}

@inproceedings{zhong25c_interspeech,
  title     = {{Pairwise Evaluation of Accent Similarity in Speech Synthesis}},
  author    = {Jinzuomu Zhong and Suyuan Liu and Dan Wells and Korin Richmond},
  year      = {2025},
  booktitle = {{Interspeech 2025}},
  pages     = {2290--2294},
  doi       = {10.21437/Interspeech.2025-2283},
  issn      = {2958-1796},
}

@article{prakash2023exploring,
  author={Prakash, Anusha and Murthy, Hema A.},
  journal={IEEE/ACM Transactions on Audio, Speech, and Language Processing}, 
  title={Exploring the Role of Language Families for Building Indic Speech Synthesisers}, 
  year={2023},
  volume={31},
  number={},
  pages={734-747},
  doi={10.1109/TASLP.2022.3230453}}

@misc{chen2025f5ttsfairytalerfakesfluent,
      title={F5-TTS: A Fairytaler that Fakes Fluent and Faithful Speech with Flow Matching}, 
      author={Yushen Chen and Zhikang Niu and Ziyang Ma and Keqi Deng and Chunhui Wang and Jian Zhao and Kai Yu and Xie Chen},
      year={2025},
      eprint={2410.06885},
      archivePrefix={arXiv},
      primaryClass={eess.AS},
      url={https://arxiv.org/abs/2410.06885}, 
}

@misc{varadhan2025phirherafairyenglish,
      title={Phir Hera Fairy: An English Fairytaler is a Strong Faker of Fluent Speech in Low-Resource Indian Languages}, 
      author={Praveen Srinivasa Varadhan and Srija Anand and Soma Siddhartha and Mitesh M. Khapra},
      year={2025},
      eprint={2505.20693},
      archivePrefix={arXiv},
      primaryClass={cs.CL},
      url={https://arxiv.org/abs/2505.20693}, 
}

@inproceedings{chen2016_xgboost,
author = {Chen, Tianqi and Guestrin, Carlos},
title = {XGBoost: A Scalable Tree Boosting System},
year = {2016},
isbn = {9781450342322},
publisher = {Association for Computing Machinery},
address = {New York, NY, USA},
url = {https://doi.org/10.1145/2939672.2939785},
doi = {10.1145/2939672.2939785},
booktitle = {Proceedings of the 22nd ACM SIGKDD International Conference on Knowledge Discovery and Data Mining},
pages = {785–794},
numpages = {10},
location = {San Francisco, California, USA},
series = {KDD '16}
}

@inproceedings{lundberg2017_shape,
author = {Lundberg, Scott M. and Lee, Su-In},
title = {A unified approach to interpreting model predictions},
year = {2017},
isbn = {9781510860964},
publisher = {Curran Associates Inc.},
address = {Red Hook, NY, USA},
booktitle = {Proceedings of the 31st International Conference on Neural Information Processing Systems},
pages = {4768–4777},
numpages = {10},
location = {Long Beach, California, USA},
series = {NIPS'17}
}

@inproceedings{cooper23_interspeech,
  title     = {{Investigating Range-Equalizing Bias in Mean Opinion Score Ratings of Synthesized Speech}},
  author    = {Erica Cooper and Junichi Yamagishi},
  year      = {2023},
  booktitle = {{Interspeech 2023}},
  pages     = {1104--1108},
  doi       = {10.21437/Interspeech.2023-1076},
  issn      = {2958-1796},
}

@article{perrotin2025_blizzard,
author = {Perrotin, Olivier and Stephenson, Brooke and Gerber, Silvain and Bailly, G\'{e}rard and King, Simon},
title = {Refining the evaluation of speech synthesis: A summary of the Blizzard Challenge 2023},
year = {2025},
issue_date = {Mar 2025},
publisher = {Academic Press Ltd.},
address = {GBR},
volume = {90},
number = {C},
issn = {0885-2308},
url = {https://doi.org/10.1016/j.csl.2024.101747},
doi = {10.1016/j.csl.2024.101747},
journal = {Comput. Speech Lang.},
month = mar,
numpages = {22}
}

@article{diciccio1996_bootstrapconfidence,
author = {Thomas J. DiCiccio and Bradley Efron},
title = {{Bootstrap confidence intervals}},
volume = {11},
journal = {Statistical Science},
number = {3},
publisher = {Institute of Mathematical Statistics},
pages = {189 -- 228},
year = {1996},
doi = {10.1214/ss/1032280214},
URL = {https://doi.org/10.1214/ss/1032280214}
}

@article{hunter2004_mmbt,
author = {David R. Hunter},
title = {{MM algorithms for generalized Bradley-Terry models}},
volume = {32},
journal = {The Annals of Statistics},
number = {1},
publisher = {Institute of Mathematical Statistics},
pages = {384 -- 406},
year = {2004},
doi = {10.1214/aos/1079120141},
URL = {https://doi.org/10.1214/aos/1079120141}
}

@inproceedings{dall14_speechprosody,
  title     = {{Rating Naturalness in Speech Synthesis: The Effect of Style and Expectation}},
  author    = {Rasmus Dall and Junichi Yamagishi and Simon King},
  year      = {2014},
  booktitle = {{Speech Prosody 2014}},
  pages     = {1012--1016},
  doi       = {10.21437/SpeechProsody.2014-192},
  issn      = {2333-2042},
}

@inproceedings{wester2015are,
  author       = {Mirjam Wester and
                  Cassia Valentini{-}Botinhao and
                  Gustav Eje Henter},
  title        = {Are we using enough listeners? no! - an empirically-supported critique
                  of interspeech 2014 {TTS} evaluations},
  booktitle    = {{INTERSPEECH} 2015, 16th Annual Conference of the International Speech
                  Communication Association, Dresden, Germany, September 6-10, 2015},
  pages        = {3476--3480},
  publisher    = {{ISCA}},
  year         = {2015},
  url          = {https://doi.org/10.21437/Interspeech.2015-689},
  doi          = {10.21437/INTERSPEECH.2015-689},
  bibsource    = {dblp computer science bibliography, https://dblp.org}
}

@inproceedings{kayyar2023subjective,
  title     = {{Subjective Evaluation of Text-to-Speech Models: Comparing Absolute Category Rating and Ranking by Elimination Tests}},
  author    = {Kishor Kayyar and Christian Dittmar and Nicola Pia and Emanuel Habets},
  year      = {2023},
  booktitle = {{12th ISCA Speech Synthesis Workshop (SSW2023)}},
  pages     = {191--196},
  doi       = {10.21437/SSW.2023-30},
}

@inproceedings{kirkland23_ssw,
  author={Ambika Kirkland and Shivam Mehta and Harm Lameris and Gustav Eje Henter and Eva Szekely and Joakim Gustafson},
  title={{Stuck in the MOS pit: A critical analysis of MOS test methodology in TTS evaluation}},
  year=2023,
  booktitle={Proc. 12th ISCA Speech Synthesis Workshop (SSW2023)},
  pages={41--47},
  doi={10.21437/SSW.2023-7}
}

@misc{chiang2023report,
      title={Why We Should Report the Details in Subjective Evaluation of TTS More Rigorously}, 
      author={Cheng-Han Chiang and Wei-Ping Huang and Hung-yi Lee},
      year={2023},
      eprint={2306.02044},
      archivePrefix={arXiv},
      primaryClass={eess.AS}
}

@misc{mushra2015,
  author       = {ITU-R},
  title        = {Method for the subjective assessment of intermediate quality level of audio systems},
  howpublished = {\url{https://www.itu.int/dms_pubrec/itu-r/rec/bs/R-REC-BS.1534-3-201510-I!!PDF-E.pdf}},
  year         = {2015}
}

@inproceedings{
shen2024naturalspeech,
title={NaturalSpeech 2: Latent Diffusion Models are Natural and Zero-Shot Speech and Singing Synthesizers},
author={Kai Shen and Zeqian Ju and Xu Tan and Eric Liu and Yichong Leng and Lei He and Tao Qin and sheng zhao and Jiang Bian},
booktitle={The Twelfth International Conference on Learning Representations},
year={2024},
url={https://openreview.net/forum?id=Rc7dAwVL3v}
}

@misc{ju2024naturalspeech,
      title={NaturalSpeech 3: Zero-Shot Speech Synthesis with Factorized Codec and Diffusion Models}, 
      author={Zeqian Ju and Yuancheng Wang and Kai Shen and Xu Tan and Detai Xin and Dongchao Yang and Yanqing Liu and Yichong Leng and Kaitao Song and Siliang Tang and Zhizheng Wu and Tao Qin and Xiang-Yang Li and Wei Ye and Shikun Zhang and Jiang Bian and Lei He and Jinyu Li and Sheng Zhao},
      year={2024},
      eprint={2403.03100},
      archivePrefix={arXiv},
      primaryClass={eess.AS}
}

@article{lemaguer2024limits,
title = {The limits of the Mean Opinion Score for speech synthesis evaluation},
journal = {Computer Speech and Language},
volume = {84},
pages = {101577},
year = {2024},
issn = {0885-2308},
doi = {https://doi.org/10.1016/j.csl.2023.101577},
url = {https://www.sciencedirect.com/science/article/pii/S0885230823000967},
author = {Sébastien {Le Maguer} and Simon King and Naomi Harte}
}

@Inbook{Loizou2011,
author="Loizou, Philipos C.",
editor="Lin, Weisi
and Tao, Dacheng
and Kacprzyk, Janusz
and Li, Zhu
and Izquierdo, Ebroul
and Wang, Haohong",
title="Speech Quality Assessment",
bookTitle="Multimedia Analysis, Processing and Communications",
year="2011",
publisher="Springer Berlin Heidelberg",
address="Berlin, Heidelberg",
pages="623--654",
isbn="978-3-642-19551-8",
doi="10.1007/978-3-642-19551-8_23",
url="https://doi.org/10.1007/978-3-642-19551-8_23"
}

\end{document}